# Compilation of Weighted Finite-State Transducers from Decision Trees

**Richard Sproat**
Bell Laboratories
700 Mountain Avenue
Murray Hill, NJ, USA
rws@bell-labs.com

**Michael Riley**
AT&T Research
600 Mountain Avenue
Murray Hill, NJ, USA
riley@research.att.com

## Abstract

We report on a method for compiling decision trees into weighted finite-state transducers. The key assumptions are that the tree predictions specify how to rewrite symbols from an input string, and the decision at each tree node is stateable in terms of regular expressions on the input string. Each leaf node can then be treated as a separate rule where the left and right contexts are constructable from the decisions made traversing the tree from the root to the leaf. These rules are compiled into transducers using the weighted rewrite-rule rule-compilation algorithm described in (Mohri and Sproat, 1996).

## 1 Introduction

Much attention has been devoted recently to methods for inferring linguistic models from data. One powerful inference method that has been used in various applications are decision trees, and in particular classification and regression trees (Breiman et al., 1984).

An increasing amount of attention has also been focussed on finite-state methods for implementing linguistic models, in particular finite-state transducers and *weighted* finite-state transducers; see (Kaplan and Kay, 1994; Pereira et al., 1994, inter alia). The reason for the renewed interest in finite-state mechanisms is clear. Finite-state machines provide a mathematically well-understood computational framework for representing a wide variety of information, both in NLP and speech processing. Lexicons, phonological rules, Hidden Markov Models, and (regular) grammars are all representable as finite-state machines, and finite-state operations such as union, intersection and composition mean that information from these various sources can be combined in useful and computationally attractive ways. The reader is referred to the above-cited papers (among others) for more extensive justification.

This paper reports on a marriage of these two strands of research in the form of an algorithm for compiling the information in decision trees into weighted finite-state transducers.[1] Given this algorithm, information inferred from data and represented in a tree can be used directly in a system that represents other information, such as lexicons or grammars, in the form of finite-state machines.

## 2 Quick Review of Tree-Based Modeling

A general introduction to classification and regression trees ('CART') including the algorithm for growing trees from data can be found in (Breiman et al., 1984). Applications of tree-based modeling to problems in speech and NLP are discussed in (Riley, 1989; Riley, 1991; Wang and Hirschberg, 1992; Magerman, 1995, inter alia). In this section we presume that one has already trained a tree or set of trees, and we merely remind the reader of the salient points in the interpretation of those trees.

Consider the tree depicted in Figure 1, which was trained on the TIMIT database (Fisher et al., 1987), and which models the phonetic realization of the English phoneme /aa/ (/ɑ/) in various environments (Riley, 1991). When this tree is used in predicting the allophonic form of a particular instance of /aa/, one starts at the root of the tree, and asks questions about the environment in which the /aa/ is found. Each non-leaf node $n$, dominates two daughter nodes conventionally labeled as $2n$ and $2n+1$; the decision on whether to go left to $2n$ or right to $2n+1$ depends on the answer to the question that is being asked at node $n$.

[1]The work reported here can thus be seen as complementary to recent reports on methods for directly inferring transducers from data (Oncina et al., 1993; Gildea and Jurafsky, 1995).

A concrete example will serve to illustrate. Consider that we have /aa/ in some environment. The first question that is asked concerns the number of segments, including the /aa/ itself, that occur to the left of the /aa/ in the word in which /aa/ occurs. (See Table 1 for an explanation of the symbols used in Figure 1.) In this case, if the /aa/ is initial — i.e., *lseg* is 1, one goes left; if there is one or more segments to the left in the word, go right. Let us assume that this /aa/ is initial in the word, in which case we go left. The next question concerns the consonantal 'place' of articulation of the segment to the right of /aa/; if it is alveolar go left; otherwise, if it is of some other quality, or if the segment to the right of /aa/ is not a consonant, then go right. Let us assume that the segment to the right is /z/, which is alveolar, so we go left. This lands us at terminal node 4. The tree in Figure 1 shows us that in the training data 119 out of 308 occurrences of /aa/ in this environment were realized as [ao], or in other words that we can estimate the probability of /aa/ being realized as [ao] in this environment as .385. The full set of realizations at this node with estimated non-zero probabilities is as follows (see Table 2 for a relevant set of ARPABET-IPA correspondences):

| phone | probability | − log prob. (weight) |
|---|---|---|
| **ao** | **0.385** | **0.95** |
| aa | 0.289 | 1.24 |
| q+aa | 0.103 | 2.27 |
| q+ao | 0.096 | 2.34 |
| ah | 0.069 | 2.68 |
| ax | 0.058 | 2.84 |

An important point to bear in mind is that a decision tree in general is a complete description, in the sense that for any new data point, there will be some leaf node that corresponds to it. So for the tree in Figure 1, each new novel instance of /aa/ will be handled by (exactly) one leaf node in the tree, depending upon the environment in which the /aa/ finds itself.

Another important point is that each decision tree considered here has the property that its predictions specify how to rewrite a symbol (in context) in an input string. In particular, they specify a two-level mapping from a set of input symbols (phonemes) to a set of output symbols (allophones).

## 3 Quick Review of Rule Compilation

Work on finite-state phonology (Johnson, 1972; Koskenniemi, 1983; Kaplan and Kay, 1994) has shown that systems of rewrite rules of the familiar form $\phi \rightarrow \psi/\lambda__\rho$, where $\phi$, $\psi$, $\lambda$ and $\rho$ are regular expressions, can be represented computationally as finite-state transducers (FSTs): note that $\phi$ represents the rule's input rule, $\psi$ the output, and $\lambda$ and $\rho$, respectively, the left and right contexts.

Kaplan and Kay (1994) have presented a concrete algorithm for compiling systems of such rules into FSTs. These methods can be extended slightly to include the compilation of *probabilistic* or *weighted* rules into weighted finite-state-transducers (WFSTs — see (Pereira et al., 1994)): Mohri and Sproat (1996) describe a rule-compilation algorithm which is more efficient than the Kaplan-Kay algorithm, and which has been extended to handle weighted rules. For present purposes it is sufficient to observe that given this extended algorithm, we can allow $\psi$ in the expression $\phi \rightarrow \psi/\lambda__\rho$, to represent a weighted regular expression. The compiled transducer corresponding to that rule will replace $\phi$ with $\psi$ with the appropriate weights in the context $\lambda__\rho$.

## 4 The Tree Compilation Algorithm

The key requirements on the kind of decision trees that we can compile into WFSTs are (1) the predictions at the leaf nodes specify how to rewrite a particular symbol in an input string, and (2) the decisions at each node are stateable as regular expressions over the input string. Each leaf node represents a single rule. The regular expressions for each branch describe one aspect of the left context $\lambda$, right context $\rho$, or both. The left and right contexts for the rule consist of the *intersections* of the partial descriptions of these contexts defined for each branch traversed between the root and leaf node. The input $\phi$ is predefined for the entire tree, whereas the output $\psi$ is defined as the union of the set of outputs, along with their weights, that are associated with the leaf node. The weighted rule belonging to the leaf node can then be compiled into a transducer using the weighted-rule-compilation algorithm referenced in the preceding section. The transducer for the entire tree can be derived by the *intersection* of the entire set of transducers associated with the leaf nodes. Note that while regular relations are not generally closed under intersection, the subset of *same-length* (or more strictly speaking *length-preserving*) relations is closed; see below.

To see how this works, let us return to the example in Figure 1. To start with, we know that this tree models the phonetic realization of /aa/, so we can immediately set $\phi$ to be $aa$ for the whole tree. Next, consider again the traversal of the tree from the root node to leaf node 4. The first decision concerns the number of segments to the left of the /aa/ in the word, either none for the left branch, or one or more for the right branch. Assuming that we have a symbol $\alpha$ representing a single segment, the symbol # representing a word boundary, and allowing for the possibility of intervening optional stress marks — $'$ — which do not count as segments, these two possibilities can be represented by the regular expressions for $\lambda$ in (a) of Table 3.[2] At this node there is no decision based on the righthand context, so the righthand context is free. We can represent this by setting $\rho$ at this node to be $\Sigma^*$, where $\Sigma$ (conventionally) represents the entire alphabet: note that the alphabet is defined to be an alphabet of all $\phi{:}\psi$ correspondence pairs that were determined empirically to be possible.

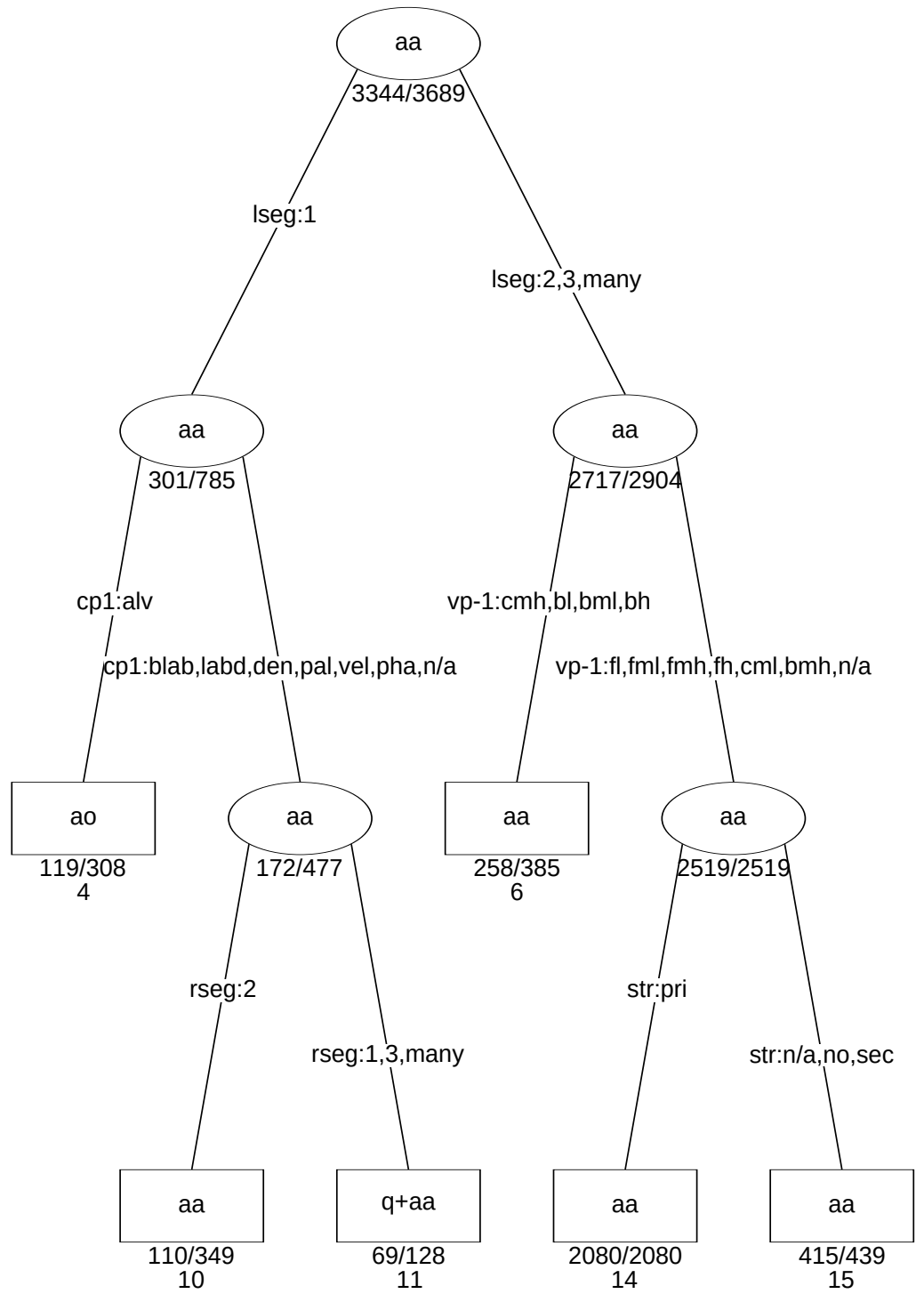

Figure 1: Tree modeling the phonetic realization of /aa/. All phones are given in ARPABET. Table 2 gives ARPABET-IPA conversions for symbols relevant to this example. See Table 1 for an explanation of other symbols

| | |
|---|---|
| *cp*n | place of articulation of consonant $n$ segments to the right |
| *cp*-n | place of articulation of consonant $n$ segments to the left |
| | values: **alv**eolar; **bilab**ial; **lab**io**d**ental; **den**tal; **pal**atal; **vel**ar; **pha**ryngeal; |
| | **n/a** if is a vowel, or there is no such segment |
| *vp*n | place of articulation of vowel $n$ segments to the right |
| *vp*-n | place of articulation of vowel $n$ segments to the left |
| | values: **c**entral-**m**id-**h**igh; **b**ack-**l**ow; **b**ack-**m**id-**l**ow; **b**ack-**h**igh; **f**ront-**l**ow; |
| | **f**ront-**m**id-**l**ow; **f**ront-**m**id-**h**igh; **f**ront-**h**igh; **c**entral-**m**id-**l**ow; **b**ack-**m**id-**h**igh |
| | **n/a** if is a consonant, or there is no such segment |
| *lseg* | number of preceding segments *including* the segment of interest *within the word* |
| *rseg* | number of following segments *including* the segment of interest *within the word* |
| | values: **1**, **2**, **3**, **many** |
| *str* | stress assigned to this vowel |
| | values: **pri**mary, **sec**ondary, **no** (zero) stress |
| | **n/a** if there is no stress mark |

Table 1: Explanation of symbols in Figure 1.

| | |
|---|---|
| aa | ɑ |
| ao | ɔ |
| ax | ə |
| ah | ʌ |
| q+aa | ʔɑ |
| q+ao | ʔɔ |

Table 2: ARPABET-IPA conversion for symbols relevant for Figure 1.

The decision at the left daughter of the root node concerns whether or not the segment to the right is an alveolar. Assuming we have defined classes of segments *alv*, *blab*, and so forth (represented as unions of segments) we can represent the regular expression for $\rho$ as in (b) of Table 3. In this case it is $\lambda$ which is unrestricted, so we can set that at $\Sigma^*$.

We can derive the $\lambda$ and $\rho$ expressions for the rule at leaf node 4 by intersecting together the expressions for these contexts defined for each branch traversed on the way to the leaf. For leaf node 4, $\lambda = \#Opt(') \cap \Sigma^* = \#Opt(')$, and $\rho = \Sigma^* \cap Opt(')(alv) = Opt(')(alv)$.[3] The rule input $\phi$ has already been given as *aa*. The output $\psi$ is defined as the union of all of the possible expressions — at the leaf node in question — that *aa* could become, with their associated weights (negative log probabilities), which we represent here as subscripted floating-point numbers:

$$\psi = ao_{0.95} \cup aa_{1.24} \cup q{+}aa_{2.27} \cup q{+}ao_{2.34} \cup ah_{2.68} \cup ax_{2.84}$$

Thus the entire weighted rule can be written as follows:

$$aa \rightarrow (ao_{0.95} \cup aa_{1.24} \cup q{+}aa_{2.27} \cup q{+}ao_{2.34} \cup ah_{2.68} \cup ax_{2.84}) / \#Opt(')___Opt(')(alv)$$

By a similar construction, the rule at node 6, for example, would be represented as:

$$aa \rightarrow (aa_{0.40} \cup ao_{1.11}) / (\#(Opt(')\alpha)^{+}Opt(')) \cap (\Sigma^*((cmh) \cup (bl) \cup (bml) \cup (bh)))___\Sigma^*$$

Each node thus represents a rule which states that a mapping occurs between the input symbol $\phi$ and the weighted expression $\psi$ in the condition described by $\lambda___\rho$. Now, in cases where $\phi$ finds itself in a context that is *not* subsumed by $\lambda___\rho$, the rule behaves exactly as a two-level *surface coercion rule* (Koskenniemi, 1983): it freely allows $\phi$ to correspond to any $\psi$ as specified by the alphabet of pairs. These $\phi{:}\psi$ correspondences are, however, constrained by *other* rules derived from the tree, as we shall see directly.

The interpretation of the full tree is that it represents the conjunction of all such mappings: for rules $1, 2 \ldots n$, $\phi$ corresponds to $\psi_1$ given condition $\lambda_1___\rho_1$ **and** $\phi$ corresponds to $\psi_2$ given condition $\lambda_2___\rho_2$ ...**and** $\phi$ corresponds to $\psi_n$ given condition $\lambda_n___\rho_n$. But this conjunction is simply the *intersection* of the entire set of transducers defined for the leaves of the tree. Observe now that the $\phi{:}\psi$ correspondences that were left free by the rule of one leaf node, are constrained by intersection with the other leaf nodes: since, as noted above, the tree is a complete description, it follows that for any leaf node $i$, and for any context $\lambda___\rho$ *not* subsumed by $\lambda_i___\rho_i$, there is some leaf node $j$ such that $\lambda_j___\rho_j$ subsumes $\lambda___\rho$. Thus, the transducers compiled for the rules at nodes 4 and 6, are intersected together, along with the rules for all the other leaf nodes. Now, as noted above, and as discussed by Kaplan and Kay (1994) regular relations — the algebraic counterpart of FSTs — are not in general closed under intersection; however, the subset of *same-length* regular relations *is* closed under intersection, since they can be thought of as finite-state *acceptors* ex-pressed over *pairs* of symbols.[4] This point can be extended somewhat to include relations that involve *bounded* deletions or insertions: this is precisely the interpretation necessary for systems of two-level rules (Koskenniemi, 1983), where a single transducer expressing the entire system may be constructed via intersection of the transducers expressing the individual rules (Kaplan and Kay, 1994, pages 367–376). Indeed, our decision tree represents neither more nor less than a set of *weighted* two-level rules. Each of the symbols in the expressions for $\lambda$ and $\rho$ actually represent (sets of) pairs of symbols: thus *alv*, for example, represents all lexical alveolars paired with all of their possible surface realizations. And just as each tree represents a system of weighted two-level rules, so a set of trees — e.g., where each tree deals with the realization of a particular phone — represents a system of weighted two-level rules, where each two-level rule is compiled from each of the individual trees.

[2] As far as possible, we use the notation of Kaplan and Kay (1994).

[3] Strictly speaking, since the $\lambda$s and $\rho$s at each branch may define expressions of different lengths, it is necessary to left-pad each $\lambda$ with $\Sigma^*$, and right-pad each $\rho$ with $\Sigma^*$. We gloss over this point here in order to make the regular expressions somewhat simpler to understand

| | | | | | |
|---|---|---|---|---|---|
| (a) | left branch | $\lambda$ | = | $\#Opt(') $ |
| | | $\rho$ | = | $\Sigma^*$ |
| | right branch | $\lambda$ | = | $(\#Opt(')\alpha Opt(')) \cup (\#Opt(')\alpha Opt(')\alpha Opt('))\cup$ $(\#Opt(')\alpha Opt(')\alpha Opt(')(\alpha Opt('))^+)$ |
| | | | = | $(Opt(')\alpha)^+ Opt(')$ |
| | | $\rho$ | = | $\Sigma^*$ |
| (b) | left branch | $\lambda$ | = | $\Sigma^*$ |
| | | $\rho$ | = | $Opt(')(alv)$ |
| | right branch | $\lambda$ | = | $\Sigma^*$ |
| | | $\rho$ | = | $(Opt(')(blab)) \cup (Opt(')(labd)) \cup (Opt(')(den)) \cup (Opt(')(pal))\cup$ $(Opt(')(vel)) \cup (Opt(')(pha)) \cup (Opt(')(n/a))$ |

Table 3: Regular-expression interpretation of the decisions involved in going from the root node to leaf node 4 in the tree in Figure 1. Note that, as per convention, superscript '+' denotes one or more instances of an expression.

We can summarize this discussion more formally as follows. We presume a function *Compile* which given a rule returns the WFST computing that rule. The WFST for a single leaf $L$ is thus defined as follows, where $\phi_T$ is the input symbol for the entire tree, $\psi_L$ is the output expression defined at $L$, $P_L$ represents the path traversed from the root node to $L$, $p$ is an individual branch on that path, and $\lambda_p$ and $\rho_p$ are the expressions for $\lambda$ and $\rho$ defined at $p$:

$$Rule_L = Compile(\phi_T \rightarrow \psi_L / \bigcap_{p \in P_L} \lambda_p __ \bigcap_{p \in P_L} \rho_p)$$

The transducer for an entire tree $T$ is defined as:

$$Rule_T = \bigcap_{L \in T} Rule_L$$

Finally, the transducer for a forest $F$ of trees is just:

$$Rule_F = \bigcap_{T \in F} Rule_T$$

## 5 Empirical Verification of the Method.

The algorithm just described has been empirically verified on the Resource Management (RM) continuous speech recognition task (Price et al., 1988). Following somewhat the discussion in (Pereira et al., 1994; Pereira and Riley, 1996), we can represent the speech recognition task as the problem of finding the best path in the composition of a grammar (language model) $G$, the transitive-closure of a dictionary $D$ mapping between words and their *phonemic* representation, a model of phone realization $\Phi$, and a weighted lattice representing the acoustic observations $A$.

[4] One can thus define intersection for transducers analogously with intersection for acceptors. Given two machines $G_1$ and $G_2$, with transition functions $\delta_1$ and $\delta_2$, one can define the transition function of $G$, $\delta$, as follows: for an input-output pair $(i, o)$, $\delta((q_1, q_2), (i, o)) = (q_1', q_2')$ if and only if $\delta_1(q_1, (i, o)) = q_1'$ and $\delta_2(q_2, (i, o)) = q_2'$.

Thus:

$$BestPath(G \circ D^* \circ \Phi \circ A) \qquad (1)$$

The transducer $\Phi = \bigcap_{T \in F} Rule_T$ can be constructed out of the forest $F$ of 40 trees, one for each phoneme, trained on the TIMIT database. The size of the trees range from 1 to 23 leaf nodes, with a total of 291 leaves for the entire forest.

The model was tested on 300 sentences from the RM task containing 2560 word tokens, and approximately 10,500 phonemes. A version of the model of recognition given in expression (1), where $\Phi$ is a transducer computed from the trees, was compared with a version where the trees were used directly following a method described in (Ljolje and Riley, 1992). The phonetic realizations and their weights were identical for both methods, thus verifying the correctness of the compilation algorithm described here.

The sizes of the compiled transducers can be quite large; in fact they were sufficiently large that instead of constructing $\Phi$ beforehand, we intersected the 40 individual transducers with the lattice $D^*$ at runtime. Table 4 gives sizes for the entire set of phone trees: tree sizes are listed in terms of number of rules (terminal nodes) and raw size in bytes; transducer sizes are listed in terms of number of states and arcs. Note that the entire alphabet comprises 215 symbol pairs. Also given in Table 4 are the compilation times for the individual trees on a Silicon Graphics R4400 machine running at 150 MHz with 1024 Mbytes of memory. The times are somewhat slow for the larger trees, but still acceptable for off-line compilation.

While the sizes of the resulting transducers seem at first glance to be unfavorable, it is important to bear in mind that size is not the only consideration in deciding upon a particular representation. WFSTs possess several nice properties that are not shared by trees, or handwritten rulesets for that matter. In particular, once compiled into a WFST, a tree can be used in the same way as a WFST derived from any other source, such as a lexicon or a language model; a compiled WFST can be used directly in a speech recognition model such as that of (Pereira and Riley, 1996) or in a speech synthesis text-analysis model such as that of (Sproat, 1996). Use of a tree directly requires a special-purpose interpreter, which is much less flexible.

It should also be borne in mind that the size explosion evident in Table 4 also characterizes rules that are compiled from hand-built rewrite rules (Kaplan and Kay, 1994; Mohri and Sproat, 1996). For example, the text-analysis ruleset for the Bell Labs German text-to-speech (TTS) system (see (Sproat, 1996; Mohri and Sproat, 1996)) contains sets of rules for the pronunciation of various orthographic symbols. The ruleset for <a>, for example, contains 25 ordered rewrite rules. Over an alphabet of 194 symbols, this compiles, using the algorithm of (Mohri and Sproat, 1996), into a transducer containing 213,408 arcs and 1,927 states. This is 72% as many arcs and 48% as many states as the transducer for /ah/ in Table 4. The size explosion is not quite as great here, but the resulting transducer is still large compared to the original rule file, which only requires 1428 bytes of storage. Again, the advantages of representing the rules as a transducer outweigh the problems of size.[5]

## 6 Future Applications

We have presented a practical algorithm for converting decision trees inferred from data into weighted finite-state transducers that directly implement the models implicit in the trees, and we have empirically verified that the algorithm is correct.

Several interesting areas of application come to mind. In addition to speech recognition, where we hope to apply the phonetic realization models described above to the much larger North American Business task (Paul and Baker, 1992), there are also applications to TTS where, for example, the decision trees for prosodic phrase-boundary prediction discussed in (Wang and Hirschberg, 1992) can be compiled into transducers and used directly in the WFST-based model of text analysis used in the multi-lingual version of the Bell Laboratories TTS system, described in (Sproat, 1995; Sproat, 1996).

## 7 Acknowledgments

The authors wish to thank Fernando Pereira, Mehryar Mohri and two anonymous referees for useful comments.

[5] Having said this, we note that obviously one would like to decrease the size of the resulting transducers if that is possible. We are currently investigating ways to avoid precompiling the transducers beforehand, but rather to construct 'on the fly', only those portions of the transducers that are necessary for a particular intersection.

| ARPABET phone | # nodes | size of tree (bytes) | # arcs | # states | time (sec) |
|---|---|---|---|---|---|
| zh | 1 | 47 | 215 | 1 | 0.3 |
| jh | 2 | 146 | 675 | 6 | 0.8 |
| aw | 2 | 149 | 1,720 | 8 | 1 |
| f | 2 | 119 | 669 | 6 | 0.9 |
| ng | 2 | 150 | 645 | 3 | 0.8 |
| oy | 2 | 159 | 1,720 | 8 | 1 |
| uh | 2 | 126 | 645 | 3 | 0.9 |
| p | 3 | 252 | 6,426 | 90 | 4 |
| ay | 3 | 228 | 4,467 | 38 | 2 |
| m | 3 | 257 | 2,711 | 27 | 1 |
| ow | 3 | 236 | 3,010 | 14 | 3 |
| sh | 3 | 230 | 694 | 8 | 1 |
| v | 3 | 230 | 685 | 8 | 1 |
| b | 4 | 354 | 3,978 | 33 | 2 |
| ch | 4 | 353 | 3,010 | 25 | 4 |
| th | 4 | 373 | 1,351 | 11 | 2 |
| dh | 5 | 496 | 1,290 | 6 | 3 |
| ey | 5 | 480 | 11,510 | 96 | 27 |
| g | 6 | 427 | 372,339 | 3,000 | 21 |
| k | 6 | 500 | 6,013 | 85 | 9 |
| aa | 6 | 693 | 18,441 | 106 | 15 |
| ah | 7 | 855 | 40,135 | 273 | 110 |
| y | 7 | 712 | 9,245 | 43 | 12 |
| ao | 8 | 1,099 | 85,439 | 841 | 21 |
| eh | 8 | 960 | 16,731 | 167 | 13 |
| er | 8 | 894 | 101,765 | 821 | 31 |
| w | 8 | 680 | 118,154 | 1,147 | 51 |
| hh | 9 | 968 | 17,459 | 160 | 10 |
| l | 9 | 947 | 320,266 | 3,152 | 31 |
| uw | 9 | 1,318 | 44,868 | 552 | 28 |
| z | 9 | 1,045 | 1,987 | 33 | 5 |
| s | 10 | 1,060 | 175,901 | 2,032 | 25 |
| ae | 11 | 1,598 | 582,445 | 4,152 | 231 |
| iy | 11 | 1,196 | 695,255 | 9,625 | 103 |
| d | 12 | 1,414 | 36,067 | 389 | 38 |
| n | 16 | 1,899 | 518,066 | 3,827 | 256 |
| r | 16 | 1,901 | 131,903 | 680 | 69 |
| ih | 17 | 2,748 | 108,970 | 669 | 71 |
| t | 22 | 2,990 | 1,542,612 | 8,382 | 628 |
| ax | 23 | 4,281 | 295,954 | 3,966 | 77 |

Table 4: Sizes of transducers corresponding to each of the individual phone trees.